\documentclass{article}

\usepackage{arxiv}

\usepackage[utf8]{inputenc}
\usepackage[T1]{fontenc}

\usepackage{amsmath}
\usepackage{amsfonts}

\usepackage{array}
\usepackage{booktabs}

\usepackage{graphicx}
\usepackage{xcolor}

\usepackage{nicefrac}
\usepackage{microtype}

\usepackage{natbib}
\usepackage{doi}
\usepackage{url}
\usepackage{hyperref}
\setcitestyle{numbers,square,comma}

\usepackage{orcidlink}
\usepackage{authblk}

\newcommand{\mname}{ODF-Net}
\newcommand{\mnameA}{ODF-Net-18}
\newcommand{\mnameB}{ODF-Net-50}
\usepackage{eso-pic}

\usepackage{xcolor}
\definecolor{rarepink}{HTML}{EF3B67}

\title{\mname: Orientation Diffusion for\\
Corruption-Robust Recognition on Degraded Media}

\author[1]{Amit Kumar Patel\,\orcidlink{0009-0006-9871-1266}}
\author[1]{Shishir Kaphle}
\author[2]{Nand Kumar Yadav}
\author[1]{Debesh Jha\,\orcidlink{0000-0002-8078-6730}}

\affil[1]{Biomedical Perception \& Intelligence Lab,
Department of Computer Science,
University of South Dakota, USA}

\affil[2]{Department of Biomedical and Translational Sciences,
Sanford School of Medicine,
University of South Dakota, USA}

\affil[ ]{\texttt{\{amit.patel,shishir.kaphle\}@coyotes.usd.edu;
\{nand.yadav,debesh.jha\}@usd.edu}}

\date{}
\begin{document}

\AddToShipoutPictureFG*{%
  \AtPageUpperLeft{%
    \put(\LenToUnit{\dimexpr\paperwidth-0.5in\relax},
        \LenToUnit{-0.35in}){%
      \makebox[0pt][r]{%
        \parbox{0.75\paperwidth}{%
          \raggedleft
          \normalsize\textit{Accepted at the 28th IEEE International
          Symposium on Multimedia (ISM 2026)}
        }%
      }%
    }%
  }%
}

\AddToShipoutPictureFG*{%
  \AtPageLowerLeft{%
    \put(\LenToUnit{0.5in},\LenToUnit{0.3in}){%
      \parbox{\dimexpr\paperwidth-1in\relax}{%
        \centering\scriptsize
        \copyright~2026 IEEE. Personal use of this material is permitted.
        Permission from IEEE must be obtained for all other uses, in any
        current or future media, including reprinting/republishing this
        material for advertising or promotional purposes, creating new
        collective works, for resale or redistribution to servers or lists,
        or reuse of any copyrighted component of this work in other works.
      }%
    }%
  }%
}

\maketitle

\begin{abstract}
Convolutional networks lean on local texture statistics, which leaves them
fragile the moment appearance shifts, even when the objects themselves are
untouched. In multimedia systems that fragility is routine rather than
hypothetical: images and video frames reach a classifier only after lossy
compression, transcoding, rescaling, and sensor noise, and every one of those
steps perturbs appearance while leaving structure intact. The usual remedies,
corruption-aware augmentation and adversarial training, need access to
corrupted data and add real training cost. We take an architectural route
instead. \mname{} pairs a truncated ResNet trunk with a structural pathway built
from a fixed orientation filter bank and a learnable diffusion encoder, and
combines the two only at the deepest stage, where that pathway and a small
fusion head stand in for the residual stage we remove. Every model is trained on
clean data alone. On CIFAR-10-C and CIFAR-100-C, \mname{} improves corruption
robustness across all four backbone and dataset pairs while using 63\% and 57\%
fewer parameters than the full-depth ResNet-18 and ResNet-50 baselines.
\mnameA{} reaches 75.97\% mean Corruption Accuracy on CIFAR-10-C, 5.81 points
above its baseline. Pixelation gains 9.52 and 10.53 points across the two backbones and is the most improved corruption on CIFAR-100-C; JPEG compression gains 2.31 and 2.93 points. Better robustness at a smaller
parameter budget, without ever showing the model a corrupted image, is an
attractive operating point for recognition on degraded media.
\end{abstract}

\section{Introduction}
\label{sec:intro}

A ResNet-18~\cite{he2016deep} that reaches 86.3\% on clean CIFAR-10~\cite{krizhevsky2009learning} falls to 52.3\% on the same
images once glass blur is applied. Nothing about the objects has changed; a car
is still shaped like a car. What changed is the appearance statistics the
network had come to depend on. This is the well-documented texture bias of
convolutional networks~\cite{geirhos2018imagenet,
hermann2020origins}, and it is
why small shifts in appearance produce disproportionate drops in
accuracy~\cite{hendrycks2019benchmarking}.

For multimedia systems these are not synthetic edge cases. Content is encoded,
transmitted at variable bitrate, decoded, and rescaled before any recognition
model sees a pixel, which makes compression artifacts, rescaling blur, and
sensor noise the ordinary operating condition rather than the exception. A model
deployed on stored or streamed media is therefore judged, every day, on a
distribution nobody trained it for. This decoupling recurs across imaging domains: appearance changes entirely
across spectra while structure persists~\cite{yadav2023tva},
and clinical imaging adds pressures of its own: rare findings~\cite{patel2026rare}
and tight compute budgets~\cite{yadav2025mlru}.

Most answers to this problem work on the data. Augmentation,
corruption-specific training, and frequency or contrastive regularization all
help~\cite{hendrycks2019augmix,chen2020simple,he2020momentum,
ranabhat2025promoting},
but they ask you to anticipate the corruptions, tune additional hyperparameters,
or carry extra training cost. They also share an assumption worth questioning:
that the fix belongs in the training distribution. It can just as well belong in
the architecture, which is where we put it.

Work on shape bias in CNNs~\cite{geirhos2018imagenet,hermann2020origins,li2023emergence,ranabhat2025promoting} suggests what such an architecture should
encode. \mname{}\footnote{Code: \url{https://github.com/amit502/odf-net}}
runs a second stream alongside the RGB backbone whose front-end is a bank of
fixed oriented derivative filters. Those filters never receive a gradient, so
they cannot drift toward the texture-selective solutions that training pressure
otherwise finds. A learnable diffusion encoder then propagates that orientation
evidence across space, closing gaps in contours that corruption has broken. The
two streams meet once, at the deepest stage.

One design decision deserves stating plainly up front. The structural stream
does not sit alongside the full backbone; it takes the place of the deepest
residual stage. \mname{} is therefore smaller than the baseline it is compared
against rather than larger, which is unusual for an architectural robustness
method and which shapes how our results should be read. We return to what that
substitution does and does not let us claim in
Subsection~\ref{sec:limit_future}. We evaluate on CIFAR-10-C and CIFAR-100-C~\cite{krizhevsky2009learning,
hendrycks2019benchmarking} with ResNet-18 and ResNet-50
backbones~\cite{he2016deep}, trained on clean data throughout.

Our contributions are as follows:
\begin{itemize}
  \item We present \mname, a dual-stream architecture that pairs a truncated
    ResNet trunk (stem through \texttt{layer3}) with a structural stream built
    from a fixed orientation filter bank and a Laplacian diffusion encoder,
    combined by a late-stage fusion head in place of the deepest residual
    stage.
  \item We show that this configuration improves mean Corruption accuracy over full-depth ResNet baselines on CIFAR-10-C and CIFAR-100-C while using 63\% and 57\% fewer parameters, trained on clean data only, with the largest gains on the degradations most characteristic of media delivery pipelines.
  
  \item We ablate fusion placement and gating across eight architectural variants, identifying late-stage fusion as the reliable design choice: removing fusion costs 3.6--4.0\,pp, while gating shifts results by less than one point either way.
\end{itemize}

\section{Related Work}
\label{sec:related_work}
\subsection{Shape Bias in Deep Vision Models}
Standard CNNs are often strongly biased toward local texture patterns rather
than global object shape~\cite{geirhos2018imagenet,hermann2020origins}. This
texture bias contrasts with human perception, which emphasizes shape and
structural cues. Geirhos et al.~\cite{geirhos2018imagenet} showed that CNNs
trained on standard recognition datasets develop a strong texture bias, and
that training on stylized datasets emphasizing global shape can shift networks
toward shape sensitivity and improve robustness.

Recent studies have explored how shape bias emerges in CNNs. Li, Wen, Li, and
Lee~\cite{li2023emergence} demonstrated that inducing sparsity in network
activations promotes shape-sensitive representations, pointing to the role of
architectural choices in shaping this bias. Ranabhat, Wang, Patel, and
Santosh~\cite{ranabhat2025promoting} introduced frequency and contrastive-based
regularization to encourage shape-consistent feature learning, improving
robustness to common corruptions.

\subsection{Data-Driven and Contrastive Approaches for Robustness}
A widely used route to robustness is through data or feature-level
interventions. Contrastive learning frameworks such as SimCLR~\cite{chen2020simple}
and MoCo~\cite{he2020momentum} enforce consistency between augmented views,
encouraging representations that generalize to structural features. AugMix and related strategies improve robustness by exposing the network to
mixtures of corrupted images during training~\cite{hendrycks2019augmix}. While
effective, these approaches generally require additional augmented or stylized
data.

\subsection{Architectural and Inductive Biases for Structure}
Architectural inductive biases offer an alternative path to robust
representations. Architectural choices alone can improve generalization under distribution
shift: multi-scale residual fusion raises cross-dataset performance on unseen
medical benchmarks without any change to the training data~\cite{srivastava2022gmsrf}.
Attention-based mechanisms and feature recalibration modules further enhance a
network's focus on relevant spatial or channel information.
Transformers~\cite{ashish2017attention} provide global context aggregation,
while Squeeze-and-Excitation (SE) networks~\cite{hu2018squeeze} and Convolutional
Block Attention Modules (CBAM)~\cite{woo2018cbam} modulate features adaptively.
Closest in spirit to our work, Takikawa, Acuna, Jampani, and
Fidler~\cite{takikawa2019gated} proposed Gated-SCNN, a dual-stream
architecture that processes shape information in a dedicated shallow stream and
gates it into the main network for semantic segmentation. Their shape stream is
task-driven and supervised with boundary labels, whereas ours relies entirely on
fixed orientation-selective filters without any shape-specific supervision.
Capsule networks model part-whole relationships to preserve pose and
shape~\cite{sabour2017dynamic,hinton2018matrix,yadav2026capsule}.

Our approach aligns with this architectural-bias paradigm but differs in two
ways. We compute multi-scale structural representations using
orientation-selective filters and diffusion blocks, and apply structural
feature injection at the high-level RGB stage through late-stage fusion. This
keeps the architecture simple while leveraging structural cues for robustness.

\subsection{Corruption Robustness and Evaluation}
Evaluating robustness to common corruptions via CIFAR-C is a standard benchmark
for generalization~\cite{hendrycks2019benchmarking,hendrycks2021many}.
Prior approaches typically rely on data augmentation or adversarial training~\cite{madry2017towards,hendrycks2019augmix}.
Our model, by contrast, is trained solely on clean data. Robustness arises from
the architecture: explicit structural representations complement high-level RGB
features, producing more stable representations without any exposure to
corrupted samples.

Our work sits between these lines of research. Three differences stand out: the orientation front-end is fixed rather than
learned, needing no shape supervision unlike Gated-SCNN~\cite{takikawa2019gated};
structure is fused rather than used to reweight, unlike SE and
CBAM~\cite{hu2018squeeze,woo2018cbam}; and the gains arrive with \emph{fewer} parameters, not more. We depend on neither augmentation nor auxiliary losses: orientation structure is encoded explicitly and merged late, and the gains arrive with fewer parameters than the baselines rather than more. That last point separates this from most architectural robustness work,
where added robustness usually means added capacity.

\section{Methodology}
\label{sec:methodology}
\begin{figure*}[t]
\centering
\includegraphics[width=\textwidth]{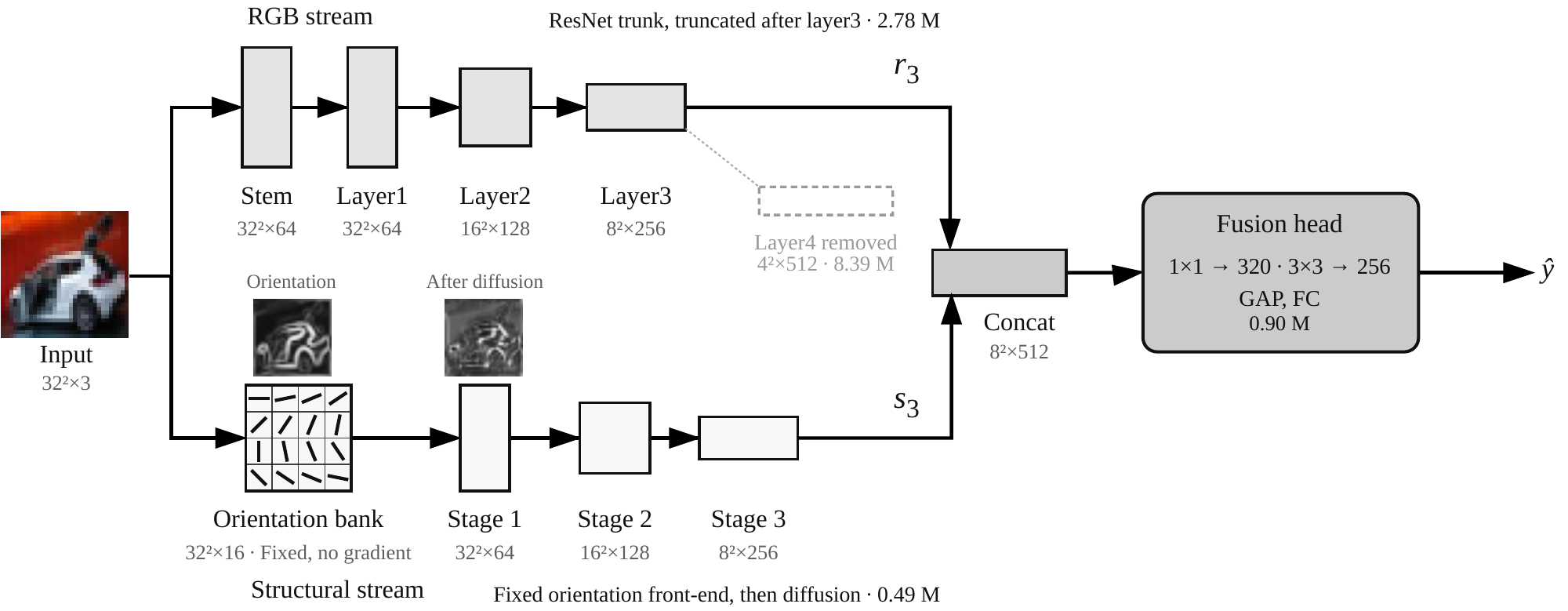}
\caption{Overview of the \mname{} architecture, shown for CIFAR inputs. The RGB
stream retains the stem and first three residual stages of a standard ResNet.
The deepest stage (\texttt{layer4}, drawn dashed) is removed, and the structural
stream and fusion head take its place; for ResNet-18 this accounts for roughly
8.4\,M of the baseline's 11.17\,M parameters, giving a 4.17\,M model overall. The
structural stream maps a grayscale projection through a fixed orientation bank,
which receives no gradient, and three stages of ShapeDiffusion blocks. The two
streams stay separate until the deepest stage, where $r_3$ and $s_3$ are
concatenated and passed to the fusion head. \mname{} therefore differs from the
baseline along two axes at once, the added structural stream and the depth of
the RGB trunk; see Subsection~\ref{sec:limit_future}.}
\label{fig:architecture}
\end{figure*}

\subsection{Overview}
\mname{} runs two streams that never talk to each other until the end. One is a
conventional RGB backbone. The other extracts orientation structure from a
grayscale view of the same image and refines it through diffusion. Keeping them
apart is deliberate: each builds its own account of the input, and the fusion
head at the deepest stage decides how to combine them.

The structural encoder and fusion head do not sit on top of a full backbone.
They occupy the slot where the deepest residual stage used to be
(Fig.~\ref{fig:architecture}), which makes \mname{} smaller than the network it
is measured against. Clean accuracy holds up under that trade, and corruption
robustness improves on every backbone and dataset pair we test.

\subsection{Structural Stream}
\subsubsection{Orientation Bank}
For an input image $x \in \mathbb{R}^{3 \times H \times W}$, we compute a
grayscale projection $\mathrm{gray}(x)$ and convolve it with a fixed bank of
oriented Sobel-type filters,
\begin{equation}
E = \{K_\theta * \mathrm{gray}(x)\}_{\theta=1}^{K},
\end{equation}
where $\{K_\theta\}$ are rotated derivative filters across $K$ orientations.
These filter responses are fixed during training and provide a stable structural
basis for the shape stream, independent of texture or color.

\subsubsection{Shape Diffusion Encoder}
The orientation responses feed into a hierarchical encoder interleaved with
ShapeDiffusion blocks. Each block propagates structural evidence across space
using a Laplacian-driven smoothing step followed by lightweight depthwise and
pointwise convolutions:
\begin{align}
X^{(t+1)} &= X^{(t)} + \lambda\, \Delta X^{(t)}, \\
Y &= \mathrm{ReLU}\!\left(\mathrm{BN}(\mathrm{PW}(\mathrm{DW}(X^{(t+1)}))) + X^{(t)}\right),
\end{align}
where $\Delta$ is a discrete Laplacian operator, DW and PW denote depthwise and
pointwise convolutions respectively, BN denotes batch normalization, and
$\lambda$ controls the diffusion strength. This mechanism produces three key
effects:
\begin{itemize}
  \item \emph{Spatial propagation}: local shape cues spread to nearby regions,
    strengthening extended contours.
  \item \emph{Fragment completion}: broken or corrupted edges are reinforced
    through neighbourhood interactions.
  \item \emph{Structural abstraction}: representations transition from
    pixel-level edges to higher-level shape organization.
\end{itemize}

The encoder produces multi-scale structural features,
\begin{equation}
S_1 = f_1(E), \quad S_2 = f_2(S_1), \quad S_3 = f_3(S_2),
\end{equation}
which grow progressively more abstract with depth.

\begin{figure*}[t]
\centering
\includegraphics[width=\textwidth]{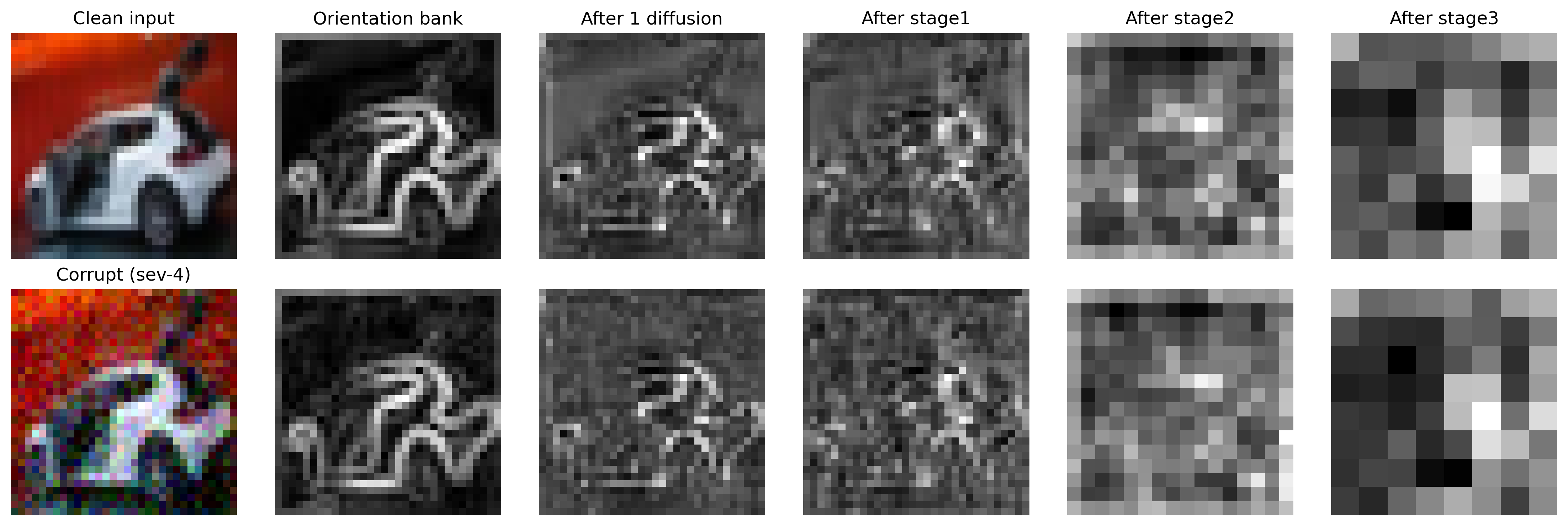}
\caption{Visualization of the structural stream for a clean image (top row) and
its corrupted counterpart (bottom row). From left to right: input image,
orientation bank output, after one ShapeDiffusion block, and after Stages 1, 2,
and 3. The orientation bank yields localized edge responses, while successive
diffusion blocks propagate and reorganize these signals into smoother, spatially
coherent representations that remain consistent under corruption.}
\label{fig:shapevis}
\end{figure*}

As shown in Fig.~\ref{fig:shapevis}, structural representations remain visually
consistent under severe corruption despite strong degradation in the RGB input.
The diffusion process does not simply extract edges; it progressively stabilizes
incomplete structural patterns by encouraging spatial continuity and suppressing
corruption-specific noise.

\subsection{RGB Backbone}
The RGB stream is a standard ResNet (ResNet-18 or ResNet-50) truncated after the
third residual stage, producing hierarchical features $R_1, R_2, R_3$. The
deepest stage (\texttt{layer4}) is removed; the structural encoder and fusion
head occupy its place in the hierarchy. While the retained features capture rich
semantic content, they remain biased toward local textures under conventional
training.

\subsection{Shape-Guided Gating}
In ablation studies, we also explore using structural features to modulate RGB
representations through a learned gating mechanism. Given RGB features $R_i$ and
aligned structural features $S_i$ at a given stage, a learned projection and
element-wise sigmoid produce a modulation mask,
\begin{align}
M_i &= \sigma(G_i(S_i)), \label{eq:mask} \\
\tilde{R}_i &= (1-\alpha)\, R_i + \alpha\, (R_i \odot M_i),
\label{eq:gate}
\end{align}
where $\sigma$ is the element-wise sigmoid, $G_i$ is a learned projection,
$\odot$ denotes element-wise multiplication, and $\alpha = \sigma(\hat{\alpha})
\in (0,1)$ is a per-gate learned scalar with raw weight $\hat{\alpha}$
initialized to 0, giving an effective starting value of $\alpha = 0.5$. We
investigate the effect of applying this mechanism at different network stages in
the ablation study; gating is absent from all main models.

\subsection{Fusion and Prediction}
The structural and RGB features are concatenated and passed through a small
fusion block consisting of $1{\times}1$ followed by $3{\times}3$ convolutions.
Global average pooling (GAP) and a linear classifier then produce the final
prediction,
\begin{equation}
\hat{y} = \mathrm{Linear}(\mathrm{GAP}(\mathrm{Fusion}([R_3, S_3]))).
\label{eq:fusion}
\end{equation}
This design integrates semantic appearance and structural information while
keeping the architecture computationally efficient.

\section{Results and Discussion}
\label{sec:results}

\subsection{Experimental Setup}
We evaluate ResNet-18 and ResNet-50 baselines against their \mname{} variants on
CIFAR-10-C and CIFAR-100-C. All models are trained exclusively on clean data for
40 epochs using stochastic gradient descent (SGD) with momentum and cosine
annealing, without any corruption augmentation, stylization, or
robustness-specific losses. Robustness is reported as mean Corruption Accuracy
(mCA) and mean Corruption Error (mCE) across 15 corruption types at 5 severity
levels~\cite{hendrycks2019benchmarking}. We define $\mathrm{mCE} = 100 -
\mathrm{mCA}$.\footnote{We report \emph{unnormalized} corruption error. This
differs from the AlexNet-normalized corruption error of Hendrycks and
Dietterich~\cite{hendrycks2019benchmarking}, and our mCE values are therefore
not directly comparable to normalized mCE reported elsewhere.} All experiments
are repeated over three random seeds and results are reported as mean $\pm$ std.
Training was conducted on a single NVIDIA RTX 5000 Ada Generation GPU (32\,GB
VRAM) with CUDA 13.2.

\begin{table}[t]
\caption{Clean accuracy and corruption robustness comparison. Bold indicates
best performance within each backbone group. Results show mean\,$\pm$\,std
across three seeds.}
\label{tab:main}
\centering
\begin{tabular}{llccc}
\toprule
\textbf{Model} & \textbf{Dataset} & \textbf{Clean Acc.} & \textbf{mCA $\uparrow$} & \textbf{mCE $\downarrow$} \\
\midrule
ResNet-18 & CIFAR-10 & 86.31$\pm$0.32 & 70.16$\pm$1.20 & 29.84$\pm$1.20 \\
\mnameA & CIFAR-10 & \textbf{86.44$\pm$0.10} & \textbf{75.97$\pm$0.23} & \textbf{24.03$\pm$0.23} \\
\midrule
ResNet-50 & CIFAR-10 & \textbf{86.26$\pm$0.76} & 69.01$\pm$0.90 & 30.99$\pm$0.90 \\
\mnameB & CIFAR-10 & 84.75$\pm$0.18 & \textbf{74.83$\pm$0.24} & \textbf{25.17$\pm$0.24} \\
\midrule
ResNet-18 & CIFAR-100 & 62.37$\pm$0.12 & 45.62$\pm$0.19 & 54.38$\pm$0.19 \\
\mnameA & CIFAR-100 & \textbf{62.46$\pm$0.19} & \textbf{47.94$\pm$0.17} & \textbf{52.06$\pm$0.17} \\
\midrule
ResNet-50 & CIFAR-100 & 59.22$\pm$1.41 & 42.51$\pm$1.00 & 57.49$\pm$1.00 \\
\mnameB & CIFAR-100 & \textbf{62.40$\pm$0.24} & \textbf{48.69$\pm$0.20} & \textbf{51.31$\pm$0.20} \\
\bottomrule
\end{tabular}
\end{table}

\subsection{Corruption Robustness}
Table~\ref{tab:main} gives the headline picture: mCA improves in all four
backbone and dataset combinations, and clean accuracy is largely preserved.

On CIFAR-10-C the two variants gain 5.81 and 5.82 percentage points over their
baselines. CIFAR-100-C is the harder setting, with ten times the classes and
correspondingly less margin, and there the gains are 2.32 points for ResNet-18
and 6.18 for ResNet-50. The ResNet-50 result on CIFAR-100 is worth pausing on:
\mnameB{} beats its baseline on \emph{clean} accuracy as well, 62.40\% against
59.22\%. A robustness intervention that also helps in-distribution is not what
one expects, and it suggests the structural stream is contributing usable signal
rather than merely regularizing. None of these numbers involved a corrupted
training image.

\begin{table*}[t]
\caption{CIFAR-10-C severity-averaged corruption accuracies (\%). Bold indicates
best accuracy within each backbone group per corruption type.}
\label{tab:cifar10_corruptions}
\centering
\begin{tabular}{lcccc}
\toprule
\textbf{Corruption} & \multicolumn{2}{c}{ResNet-18} & \multicolumn{2}{c}{ResNet-50} \\
\cmidrule(lr){2-3} \cmidrule(lr){4-5}
 & \textit{Baseline} & \textit{\mname} & \textit{Baseline} & \textit{\mname} \\
\midrule
Gaussian noise & 62.39$\pm$3.75 & \textbf{72.95$\pm$1.07} & 59.16$\pm$2.48 & \textbf{75.35$\pm$0.55} \\
Shot noise & 68.11$\pm$3.23 & \textbf{76.47$\pm$0.77} & 65.81$\pm$1.70 & \textbf{78.03$\pm$0.41} \\
Impulse noise & 61.94$\pm$2.31 & \textbf{69.02$\pm$0.74} & 62.72$\pm$0.64 & \textbf{70.76$\pm$0.66} \\
Defocus blur & 75.33$\pm$0.12 & \textbf{79.30$\pm$1.10} & 74.36$\pm$1.09 & \textbf{78.71$\pm$0.65} \\
Glass blur & 52.32$\pm$3.03 & \textbf{69.69$\pm$0.50} & 50.36$\pm$3.21 & \textbf{69.36$\pm$0.43} \\
Motion blur & 67.39$\pm$0.23 & \textbf{73.69$\pm$1.55} & 66.81$\pm$1.06 & \textbf{72.13$\pm$0.09} \\
Zoom blur & 70.09$\pm$0.52 & \textbf{76.31$\pm$1.36} & 69.14$\pm$1.62 & \textbf{75.85$\pm$0.66} \\
Snow & 72.71$\pm$1.55 & \textbf{77.18$\pm$0.31} & 70.91$\pm$0.34 & \textbf{75.14$\pm$0.08} \\
Frost & 71.28$\pm$1.83 & \textbf{76.95$\pm$0.45} & 69.61$\pm$0.57 & \textbf{73.17$\pm$0.29} \\
Fog & \textbf{75.60$\pm$0.62} & 71.96$\pm$0.51 & \textbf{75.45$\pm$1.72} & 66.93$\pm$1.00 \\
Brightness & \textbf{84.18$\pm$0.19} & 84.17$\pm$0.20 & \textbf{83.62$\pm$1.20} & 81.38$\pm$0.23 \\
Contrast & 64.12$\pm$0.81 & \textbf{71.20$\pm$0.92} & 63.00$\pm$2.20 & \textbf{66.69$\pm$1.55} \\
Elastic & 75.27$\pm$0.18 & \textbf{77.18$\pm$0.80} & 74.52$\pm$0.72 & \textbf{75.76$\pm$0.62} \\
Pixelate & 72.52$\pm$2.41 & \textbf{82.04$\pm$1.00} & 71.63$\pm$2.20 & \textbf{82.16$\pm$0.21} \\
JPEG & 79.11$\pm$0.96 & \textbf{81.42$\pm$0.27} & 78.03$\pm$0.30 & \textbf{80.96$\pm$0.21} \\
\bottomrule
\end{tabular}
\end{table*}

\begin{table*}[t]
\caption{CIFAR-100-C severity-averaged corruption accuracies (\%). Bold
indicates best accuracy within each backbone group per corruption type.}
\label{tab:cifar100_corruptions}
\centering
\begin{tabular}{lcccc}
\toprule
\textbf{Corruption} & \multicolumn{2}{c}{ResNet-18} & \multicolumn{2}{c}{ResNet-50} \\
\cmidrule(lr){2-3} \cmidrule(lr){4-5}
 & \textit{Baseline} & \textit{\mname} & \textit{Baseline} & \textit{\mname} \\
\midrule
Gaussian noise & 34.31$\pm$1.41 & \textbf{37.40$\pm$1.20} & 32.36$\pm$3.01 & \textbf{42.36$\pm$0.86} \\
Shot noise & 40.17$\pm$1.09 & \textbf{43.09$\pm$1.10} & 38.09$\pm$2.86 & \textbf{47.66$\pm$0.79} \\
Impulse noise & \textbf{33.89$\pm$1.00} & 31.79$\pm$0.97 & 33.18$\pm$1.49 & \textbf{36.98$\pm$0.53} \\
Defocus blur & 52.15$\pm$0.26 & \textbf{55.36$\pm$0.29} & 48.30$\pm$0.84 & \textbf{54.99$\pm$0.36} \\
Glass blur & 34.94$\pm$0.70 & \textbf{40.08$\pm$0.75} & 34.66$\pm$5.98 & \textbf{42.57$\pm$0.76} \\
Motion blur & 46.76$\pm$0.34 & \textbf{50.47$\pm$0.08} & 43.50$\pm$0.58 & \textbf{50.50$\pm$0.52} \\
Zoom blur & 47.91$\pm$0.47 & \textbf{53.06$\pm$0.15} & 44.29$\pm$1.05 & \textbf{52.66$\pm$0.35} \\
Snow & 48.09$\pm$0.83 & \textbf{50.08$\pm$0.41} & 44.80$\pm$0.42 & \textbf{50.51$\pm$0.14} \\
Frost & 46.93$\pm$0.74 & \textbf{49.36$\pm$0.21} & 41.74$\pm$0.68 & \textbf{48.09$\pm$0.25} \\
Fog & \textbf{49.13$\pm$0.48} & 45.40$\pm$0.24 & 43.08$\pm$1.99 & \textbf{43.37$\pm$0.48} \\
Brightness & \textbf{58.13$\pm$0.46} & 58.11$\pm$0.28 & 53.90$\pm$1.91 & \textbf{57.40$\pm$0.18} \\
Contrast & 42.23$\pm$0.96 & \textbf{43.18$\pm$0.90} & 35.53$\pm$2.03 & \textbf{39.66$\pm$0.56} \\
Elastic & 50.33$\pm$0.35 & \textbf{51.92$\pm$0.19} & 46.72$\pm$0.42 & \textbf{52.04$\pm$0.51} \\
Pixelate & 47.77$\pm$0.70 & \textbf{58.20$\pm$0.22} & 47.94$\pm$5.17 & \textbf{58.37$\pm$0.25} \\
JPEG & 51.55$\pm$0.31 & \textbf{51.68$\pm$0.44} & 49.53$\pm$0.99 & \textbf{53.14$\pm$0.23} \\
\bottomrule
\end{tabular}
\end{table*}

\begin{figure}[t]
\centering
\includegraphics[width=\linewidth]{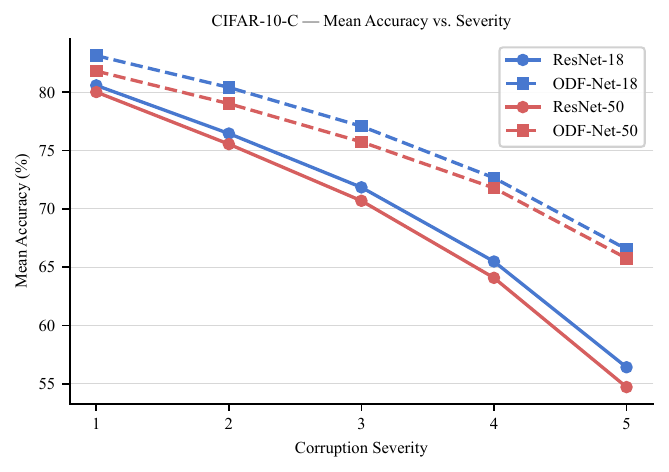}
\caption{CIFAR-10-C mean accuracy per severity level, averaged across all 15
corruption types. \mname{} models outperform their baselines as
severity increases.}
\label{fig:sev10}
\end{figure}

\begin{figure}[t]
\centering
\includegraphics[width=\linewidth]{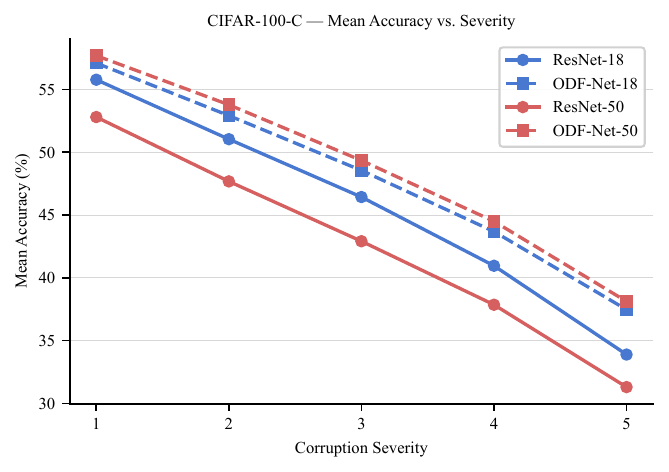}
\caption{CIFAR-100-C mean accuracy per severity level, averaged across all 15
corruption types. \mname{} models maintain higher accuracy across all severity
levels.}
\label{fig:sev100}
\end{figure}

\subsection{Per-Corruption Analysis}
Tables~\ref{tab:cifar10_corruptions} and~\ref{tab:cifar100_corruptions} report
severity-averaged accuracy for each of the 15 corruption types. The gains are not spread evenly, and where they concentrate is informative. Noise (Gaussian, shot, impulse) and blur (defocus, glass, motion, zoom) improve most, which is where texture-based representations are weakest. Glass blur is the extreme case: 52.32\% to 69.69\% on CIFAR-10-C, a 17-point swing on the
corruption that scrambles local statistics most violently while leaving coarse shape recoverable.

The corruptions tied most directly to media delivery diverge. Pixelation gains
$+9.52$ and $+10.53$\,pp on CIFAR-10-C and is the most improved corruption on
CIFAR-100-C; JPEG gains $+2.31$ and $+2.93$\,pp, improving in all four settings. This matters for the deployment setting we care about, since these are not incidental corruptions but
the ones a codec produces by design.

The exceptions are as telling as the gains. Brightness is flat for ResNet-18 and mixed for ResNet-50, fog favours the baseline in three of four settings, and impulse noise falls on CIFAR-100-C
with ResNet-18. Brightness and fog are largely low-frequency, global shifts that leave local texture legible, so a texture-reliant model has lost nothing and a structural prior has nothing to recover. The method helps
precisely where structure survives and appearance does not, which is what a shape-oriented representation ought to do.

\subsection{Efficiency}
\label{sec:efficiency}
Table~\ref{tab:efficiency} compares parameter count, compute, and runtime
characteristics. \mname{} models reduce parameter count by 63\% and 57\% for
ResNet-18 and ResNet-50 respectively, by
replacing \texttt{layer4} with the lightweight structural encoder. A ShapeDiffusion block costs $C^{2}HW+18CHW$ multiply-accumulates against
$18C^{2}HW$ for a ResNet basic block, a $14$--$17\times$ reduction at our channel
widths. The structural stream and fusion head hold 1.39\,M parameters against
\texttt{layer4}'s 8.39\,M, yet GFLOPs are nearly unchanged (0.56 vs.\ 0.55), and
latency rises because the stream adds roughly 23 layers at batch size one. The parallel stream raises latency and activation memory for
\mnameA{} but leaves \mnameB{} unaffected; the advantage is in
parameter count, not wall-clock speed.

This parameter reduction is relevant for media processing at the edge, where
recognition often runs on the same constrained device that performs decoding. A
model that is both smaller than the baseline and more tolerant of decode-side
artifacts is a favourable operating point for on-device content analysis.

\begin{table}[t]
\caption{Efficiency comparison of baseline and \mname{} models on CIFAR-10
($32{\times}32$ input, batch size 1, NVIDIA RTX 5000 Ada). Bold indicates the
more efficient value within each backbone pair.}
\label{tab:efficiency}
\centering
\setlength{\tabcolsep}{4pt}
\begin{tabular}{lccccc}
\toprule
\textbf{Model} & \textbf{Params} & \textbf{GFLOPs} & \textbf{Lat.} & \textbf{Thr.} & \textbf{Mem.} \\
 & (M) & & (ms) & (img/s) & (MB) \\
\midrule
ResNet-18 & 11.17 & 0.56 & \textbf{1.16} & \textbf{859.10} & \textbf{60.25} \\
\mnameA & \textbf{4.17} & \textbf{0.55} & 1.89 & 529.50 & 152.58 \\
\midrule
ResNet-50 & 23.52 & 1.30 & \textbf{3.30} & \textbf{302.70} & 228.22 \\
\mnameB & \textbf{10.17} & \textbf{1.18} & 3.31 & 302.10 & \textbf{177.23} \\
\bottomrule
\end{tabular}
\end{table}

\subsection{Ablations}
\label{sec:ablations}

We ablate fusion placement and gating on CIFAR-10-C with ResNet-18, holding all
training settings fixed. Results are reported in Table~\ref{tab:ablation}. All
variants share the same truncated RGB trunk, so the comparisons within this
table isolate fusion and gating choices.

\begin{table}[t]
\caption{Ablation study on CIFAR-10-C using a ResNet-18 backbone trained on
clean data only. Bold indicates best performance within each variant group.
Results show mean\,$\pm$\,std across three seeds.}
\label{tab:ablation}
\centering
\begin{tabular}{lccc}
\toprule
\textbf{Variant} & \textbf{Clean Acc.} & \textbf{mCA $\uparrow$} & \textbf{mCE $\downarrow$} \\
\midrule
ResNet-18 (baseline) & 86.31$\pm$0.32 & 70.16$\pm$1.20 & 29.84$\pm$1.20 \\
\mnameA{} (full model) & \textbf{86.44$\pm$0.10} & \textbf{75.97$\pm$0.23} & \textbf{24.03$\pm$0.23} \\
\midrule
Early gate ($r_1, r_2$) + late fusion & 86.46$\pm$0.11 & 75.60$\pm$0.82 & 24.40$\pm$0.82 \\
Late gate ($r_3$) + late fusion & 86.45$\pm$0.18 & \textbf{76.56$\pm$0.28} & \textbf{23.44$\pm$0.28} \\
Late gate ($r_3$), no fusion & 86.71$\pm$0.18 & 72.01$\pm$0.70 & 27.99$\pm$0.70 \\
Early gate, no fusion & \textbf{87.38$\pm$0.08} & 72.37$\pm$0.87 & 27.63$\pm$0.87 \\
\midrule
Early fusion ($s_2 \to r_2$) & 87.33$\pm$0.18 & 74.59$\pm$1.27 & 25.41$\pm$1.27 \\
Early fusion + early gate & \textbf{87.46$\pm$0.20} & \textbf{75.71$\pm$0.33} & \textbf{24.29$\pm$0.33} \\
Early fusion + late gate & 87.28$\pm$0.02 & 75.56$\pm$0.58 & 24.44$\pm$0.58 \\
\bottomrule
\end{tabular}
\end{table}

\subsubsection{Late fusion is the dominant contributor}
The main model, late fusion with no gating, takes mCA from 70.16\% to 75.97\%
($+5.81$\,pp). Among the choices we vary, where the streams meet matters more
than anything else. Fusing early costs more than a point (74.59\% against
75.97\%), which fits the idea that structural signal injected before the RGB
stream has formed stable low-level features interferes with that formation
rather than helping it.

\subsubsection{Fusion is necessary; gating alone is not sufficient}
Strip out fusion and keep only shape-guided gating, and performance falls to
72.01\%, below every fusion variant. Modulating RGB features with structural
information is not a substitute for actually carrying that information forward.
The structural representation has to reach the classifier, not merely reweight
what the RGB stream already computed.

\subsubsection{Gating provides marginal and inconsistent benefit}
Adding gating on top of late fusion yields results close to the main model: late
gate achieves 76.56\% ($+0.59$\,pp) while early gate achieves 75.60\%
($-0.37$\,pp). Gating therefore moves results in \emph{both} directions relative
to the gating-free model: late gating adds about twice the seed standard deviation, while early gating falls within it. The added complexity of a learned projection, batch
normalization, and scalar gate per stage does not justify this inconsistent
effect, so the main model retains the simpler gating-free design.

\subsubsection{Summary}
Fusion placement matters more than gating. Late-stage structural fusion is the
key design choice among those ablated here.

\subsection{Discussion \& Analysis }
\label{sec:disc_analysis}
\begin{figure}[t]
\centering
\includegraphics[width=\linewidth]{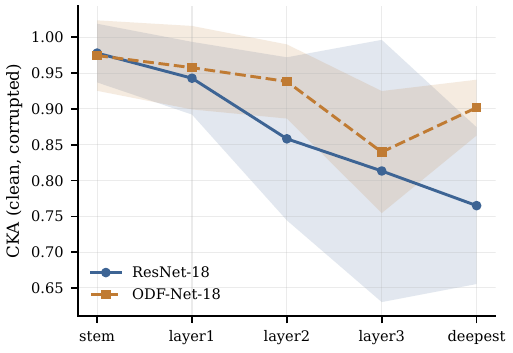}
\caption{Linear CKA between clean and corrupted representations for ResNet-18
and \mnameA{} on CIFAR-10-C at severity 3, averaged over all 15 corruption
types; the band shows one standard deviation across corruptions. Taps are
matched by role rather than by name: \emph{deepest} denotes \texttt{layer4} for
the baseline and the fusion head for \mnameA, these being the last spatial
representation each model passes to its classifier. \mnameA{} retains higher
clean-to-corrupt similarity at the deeper taps, indicating representations that
move less under corruption.}
\label{fig:cka}
\end{figure}

The corruptions that improve most are the ones that wreck local appearance while
leaving coarse spatial structure recoverable, which is the signature one would
expect if the model were leaning less on texture. Representational similarity
supports that reading directly. Linear CKA between clean and corrupted features
(Fig.~\ref{fig:cka}) stays higher for \mname{} than for the baseline, and the
gap is largest at the deepest tap, so the two models are not merely scoring
differently but representing corrupted input differently, and increasingly so
with depth. The ablation points the same way: structure helps most once the RGB
stream has built something semantic to attach it to.

Figs.~\ref{fig:sev10} and~\ref{fig:sev100} break this down by severity level.
The margin between \mname{} and the corresponding baseline widens as severity
increases on both datasets, which is the behaviour that matters for media
pipelines: the advantage is largest precisely where degradation is most
aggressive, such as low-bitrate encoding or heavy downscaling.

\subsection{Limitations \& Future Work}
\label{sec:limit_future}

\mname{} replaces \texttt{layer4} with the structural encoder and fusion head,
so our comparison against full-depth ResNet baselines varies both the structural
stream and the depth of the RGB trunk. We do not isolate the contribution of the
truncation itself, and an RGB-only ablation using the same truncated trunk and
fusion head is needed to attribute the gain to structural modeling alone.
Accordingly, we frame our result as a property of the \mname{} configuration
rather than of the structural stream in isolation.

Our evaluation is limited to CIFAR-10-C and CIFAR-100-C, conducted over three
seeds, which gives only a coarse variance estimate. Scalability to
higher-resolution benchmarks, to video, and to artifacts produced by real codecs
rather than synthetic approximations remains an open question, as structural
representations may interact differently with richer textures, greater object
diversity, and temporally correlated compression noise. Additionally, while
mean robustness improves in every setting, a gap between clean and corrupted accuracy
persists, and closing it without corruption-based augmentation remains a key
challenge.

The natural next step for a multimedia setting is temporal. Static corruption
benchmarks say nothing about whether predictions stay stable across consecutive
frames, and prediction flicker under perturbation sequences is closer to what a
deployed video pipeline actually cares about. Richer structural encoders, more
expressive orientation operators, and interaction modules beyond concatenation
are all open on the architectural side. The same argument should transfer to clinical imaging, where acquisition
differs between scanners and sites while anatomy does not, and where the
findings that matter are too rare to augment for~\cite{patel2026rare,yadav2025mlru}.

\section{Conclusion}
\label{sec:conclusion}

Our \mname{} puts an orientation-driven, diffusion-based structural stream where a
ResNet's deepest residual stage used to be, and merges it with the RGB pathway
through a small fusion head. Trained on clean data alone, it improves corruption
robustness on CIFAR-10-C and CIFAR-100-C for both a shallow and a deep backbone,
largely holds clean accuracy, and does so with 63\% and 57\% fewer parameters than the
networks it is compared against. The largest margins fall on the noise, blur,
and pixelation degradations that real delivery pipelines produce, and our
ablation shows that carrying structural features to the classifier matters more than how they are gated. Threats to validity: mCE is unnormalized, CKA is aggregate-only and single-seed,
and the depth confound is noted in Subsection~\ref{sec:limit_future}.
Robustness bought with structure rather than with data, at a smaller parameter
budget, is worth pushing further in delivery pipelines, and beyond them to
settings like clinical imaging, where acquisition changes but anatomy does not.

\bibliographystyle{plainnat}
\bibliography{main}

\section{Appendix}
\subsection{Training Protocol}
All models are trained from scratch on clean data only, with no corruption-based
augmentation, adversarial training, or robustness-specific losses. Inputs are
normalized using per-dataset mean and standard deviation. All experiments were
run on a single NVIDIA RTX 5000 Ada Generation GPU (32\,GB VRAM, CUDA 13.2). Key
hyperparameters are listed below.
\begin{itemize}
  \item \emph{Optimizer}: SGD with Nesterov momentum 0.9, weight decay
    $10^{-4}$, initial learning rate 0.1, cosine annealing over the full
    40-epoch training horizon ($T_{\max}=40$, minimum learning rate 0).
  \item \emph{Loss}: Cross-entropy with label smoothing 0.1; gradient clipping
    at $\ell_2$ norm 5.0.
  \item \emph{CIFAR}: 40 epochs, batch size 128, no crop or flip augmentation. Best checkpoint by clean test accuracy, identically for all models.
  \item \emph{Seeds}: Three seeds (42, 17, 23); results reported as mean $\pm$
    std.
  \item \emph{Gating (ablation only)}: The influence parameter $\alpha =
    \sigma(\hat{\alpha})$ is a learned scalar per gate, with raw weight
    $\hat{\alpha}$ initialized to 0, giving an effective starting value of
    $\alpha = 0.5$. No warm-up or schedule is applied.
\end{itemize}

\subsection{Architecture Details}
\textbf{Orientation Bank.} The input $x \in \mathbb{R}^{3 \times H \times W}$ is
projected to grayscale by channel averaging: $x_g = \frac{1}{3}\sum_{c=1}^{3}
x_c$. For $K=16$ orientations $\theta_k = \pi k / K$, $k = 0, \dots, K-1$,
oriented Sobel filters are constructed as $G_{\theta_k} = \cos(\theta_k) G_x +
\sin(\theta_k) G_y$, where
\begin{equation*}
G_x = \begin{bmatrix} 1 & 0 & -1 \\ 2 & 0 & -2 \\ 1 & 0 & -1 \end{bmatrix},
\quad
G_y = \begin{bmatrix} 1 & 2 & 1 \\ 0 & 0 & 0 \\ -1 & -2 & -1 \end{bmatrix}.
\end{equation*}
Each response $e_k = |x_g * G_{\theta_k}|$ is normalized by its spatial mean:
$\hat{e}_k = e_k / (\varepsilon + \mathrm{mean}_{h,w}(e_k))$, with $\varepsilon
= 10^{-6}$. This produces a stable 16-channel orientation tensor independent of
color and illumination.

\textbf{Shape Diffusion Block.} Each block applies a fixed discrete Laplacian
update $X' = X + \lambda \Delta X$, where $\Delta X = X * L$ with
\begin{equation*}
L = \begin{bmatrix} 0 & 1 & 0 \\ 1 & -4 & 1 \\ 0 & 1 & 0 \end{bmatrix},
\quad \lambda = 0.12.
\end{equation*}
The updated map is then refined via depthwise $3{\times}3$ and pointwise
$1{\times}1$ convolutions with batch normalization and a residual connection:
$Y = \mathrm{ReLU}(\mathrm{BN}(\mathrm{PW}(\mathrm{DW}(X'))) + X)$. The choice
$\lambda = 0.12$ satisfies the von Neumann stability bound $\lambda \le 0.25$,
ensuring the diffusion step does not amplify any frequency component.

\textbf{Structural Encoder.} Three hierarchical stages produce multi-scale
features ($s_1, s_2, s_3$):
\begin{itemize}
  \item Stage 1: $1{\times}1$ conv ($16 \to 64$), BN, ReLU, two diffusion
    blocks. Spatial resolution preserved.
  \item Stage 2: $3{\times}3$ conv ($64 \to 128$, stride 2), BN, ReLU, two
    diffusion blocks.
  \item Stage 3: $3{\times}3$ conv ($128 \to 256$, stride 2), BN, ReLU, one
    diffusion block.
\end{itemize}
For CIFAR ($32{\times}32$ input), spatial sizes are $s_1$: $32{\times}32$,
$s_2$: $16{\times}16$, $s_3$: $8{\times}8$. Only $s_3$ is used for fusion.

\textbf{RGB Backbone}
\mnameA{} and \mnameB{} use the stem and first three residual stages
(\texttt{layer1} through \texttt{layer3}) of the respective ResNets, producing
features ($r_1, r_2, r_3$). \texttt{layer4} is excluded; fusion occurs at
\texttt{layer3}. For ResNet-18, \texttt{layer4} accounts for approximately
8.4\,M of the 11.17\,M baseline parameters, so the retained RGB trunk contains
roughly 2.8\,M parameters and the structural stream and fusion head account for
the remainder of the 4.17\,M total. For CIFAR inputs, the standard $7{\times}7$
stem is replaced by a single $3{\times}3$ convolution (stride 1, padding 1)
followed by BN and ReLU, preserving the $32{\times}32$ resolution. No pretrained
weights are used.

\textbf{Fusion Head and Gating}
$r_3$ and $s_3$ are concatenated channel-wise and passed through: $1{\times}1$
conv ($r_3$-ch $+\,256 \to 320$), BN, ReLU, $3{\times}3$ conv ($320 \to 256$),
BN, ReLU, global average pooling, and a linear classifier (see
\eqref{eq:fusion}). For gating ablations, structural features modulate RGB
features as in \eqref{eq:mask} and \eqref{eq:gate}, where $\alpha = \sigma(\hat{\alpha})$ is a learned
scalar initialized at 0.5; gating is absent from all main models.
\end{document}